\documentclass[pdflatex,sn-mathphys-num]{sn-jnl}

\usepackage{graphicx}%
\usepackage{multirow}%
\usepackage{amsmath,amssymb,amsfonts}%
\usepackage{amsthm}%
\usepackage{mathrsfs}%
\usepackage[title]{appendix}%
\usepackage[table,xcdraw]{xcolor}%
\usepackage{textcomp}%
\usepackage{manyfoot}%
\usepackage{booktabs}%
\usepackage{algorithm}%
\usepackage{algorithmicx}%
\usepackage{algpseudocode}%
\usepackage{listings}%
\usepackage{setspace}%
\usepackage{array,hhline}
\usepackage{tabularray} 
\usepackage{rotating}
\usepackage[justification=centering]{caption}

\usepackage[T1]{fontenc}
\UseRawInputEncoding

\DefTblrTemplate{caption-tag}{default}{\textbf{Table\hspace{0.25em}\thetable}}
\DefTblrTemplate{caption-sep}{default}{\enskip}

\newcommand\rot[1]{\rotatebox{90}{#1}}

\ExplSyntaxOn
\cs_new_protected:Npn \mysetcellcolor #1
  {
    \bool_lazy_and:nnT
      { \lTblrMeasuringBool }
      { ! \tl_if_empty_p:n {#1} }
      {
        \__tblr_cell_gput:ne { background } {black!\fp_to_tl:n{#1*100}!white}
        \fp_compare:nTF {#1 > 0.3}
          {
            \__tblr_cell_gput:ne { foreground } {white}
          }
          {
            \__tblr_cell_gput:ne { foreground } {black}
          }
      }
    #1
  }
\ExplSyntaxOff

\theoremstyle{thmstyleone}%
\theoremstyle{thmstyletwo}%

\theoremstyle{thmstylethree}%

\usepackage{floatpag}

\begin{document}

\title[Article Title]{Detecting Fallacies in Climate Misinformation: A Technocognitive Approach to Identifying Misleading Argumentation}


\author{\fnm{Francisco} \sur{Zanartu}}\email{francisco.zanartu@unimelb.edu.au}

\author*[2]{\fnm{John} \sur{Cook}}\email{jocook@unimelb.edu.au}

\author[1]{\fnm{Markus} \sur{Wagner}}\email{markus.wagner@monash.edu}
\author[1]{\fnm{Julian} \sur{Garcia}}\email{julian.garcia@monash.edu}

\affil[1]{\orgdiv{Department of Data Science \& AI}, \orgname{Monash University}, \orgaddress{\city{Clayton}, \postcode{3800}, \state{Victoria}, \country{Australia}}}

\affil[2]{\orgdiv{Melbourne Centre for Behaviour Change}, \orgname{University of Melbourne}, \orgaddress{\city{Parkville}, \state{Victoria}, \country{Australia}}}



\abstract{Misinformation about climate change is a complex societal issue requiring holistic, interdisciplinary solutions at the intersection between technology and psychology. One proposed solution is a ``technocognitive'' approach, involving the synthesis of psychological and computer science research. Psychological research has identified that interventions in response to misinformation require both fact-based (e.g., factual explanations) and technique-based (e.g., explanations of misleading techniques) content. However, little progress has been made on documenting and detecting fallacies in climate misinformation. In this study, we apply a previously developed critical thinking methodology for deconstructing climate misinformation, in order to develop a dataset mapping different types of climate misinformation to reasoning fallacies. This dataset is used to train a model to detect fallacies in climate misinformation. Our study shows $\text{F}_\text{1}$ scores that are 2.5 to 3.5 better than previous works. The fallacies that are easiest to detect include fake experts and anecdotal arguments, while fallacies that require background knowledge, such as oversimplification, misrepresentation, and slothful induction, are relatively more difficult to detect. This research lays the groundwork for development of solutions where automatically detected climate misinformation can be countered with generative technique-based corrections.}


\maketitle



\sloppy

\section{Introduction}\label{sec1}

Misinformation about climate change reduces climate literacy and support for policies that mitigate climate impacts~\citep{ranney2016climate} while exacerbating public polarization~\citep{cook2017neutralizing}. Efforts to communicate the reality of climate change can be cancelled out by misinformation~\citep{van2017inoculating} and ignorance about the strong degree of public acceptance about the reality of climate change is associated with ``climate silence''~\citep{geiger2016climate}. These impacts necessitate interventions that neutralize their negative influence.

A growing body of psychological research has tested a variety of interventions aimed at reducing the impact of misinformation \citep{kozyreva2022toolbox}. Two leading communication approaches are fact-based and technique-based. Fact-based corrections---also described as topic-based~\citep{schmid2019effective}---involve exposing how misinformation is false through factual explanations. Technique-based corrections---also described as logic-based~\citep{banas2013inducing,vraga2020testing}---involve explaining misleading rhetorical techniques and logical fallacies used in misinformation. \citet{schmid2019effective} found that both fact-based and technique-based corrections were effective in countering misinformation. However, \citet{vraga2020testing} found that technique-based corrections outperformed fact-based corrections as they were equally effective whether the correction was encountered before or after the misinformation, while fact-based corrections were ineffective if misinformation was shown afterwards, leading to a cancelling out effect. This result is consistent with other studies finding that factual explanations can be cancelled out if encountered alongside contradicting misinformation~\citep{cook2017neutralizing,mccright2016examining,van2017inoculating}. Technique-based interventions can also address misinformation techniques such as paltering or cherry picking which use factual statements to mislead by withholding relevant information \citep{lewandowsky2017letting}. Synthesising the body of psychological research on countering misinformation, the recommended structure of an effective debunking contains both a fact-based element explaining the facts relevant to the misinforming argument and a technique-based element explaining the misleading rhetorical techniques or logical fallacies found in the misinforming argument~\citep{lewandowsky2020debunking}.

\begin{figure}
\caption{\label{fig:FLICC}FLICC taxonomy of misinformation techniques and logical fallacies~\citep{cook2020deconstructing}.}
\centering
\includegraphics[trim={0 0 120 0},clip,width=1\textwidth]{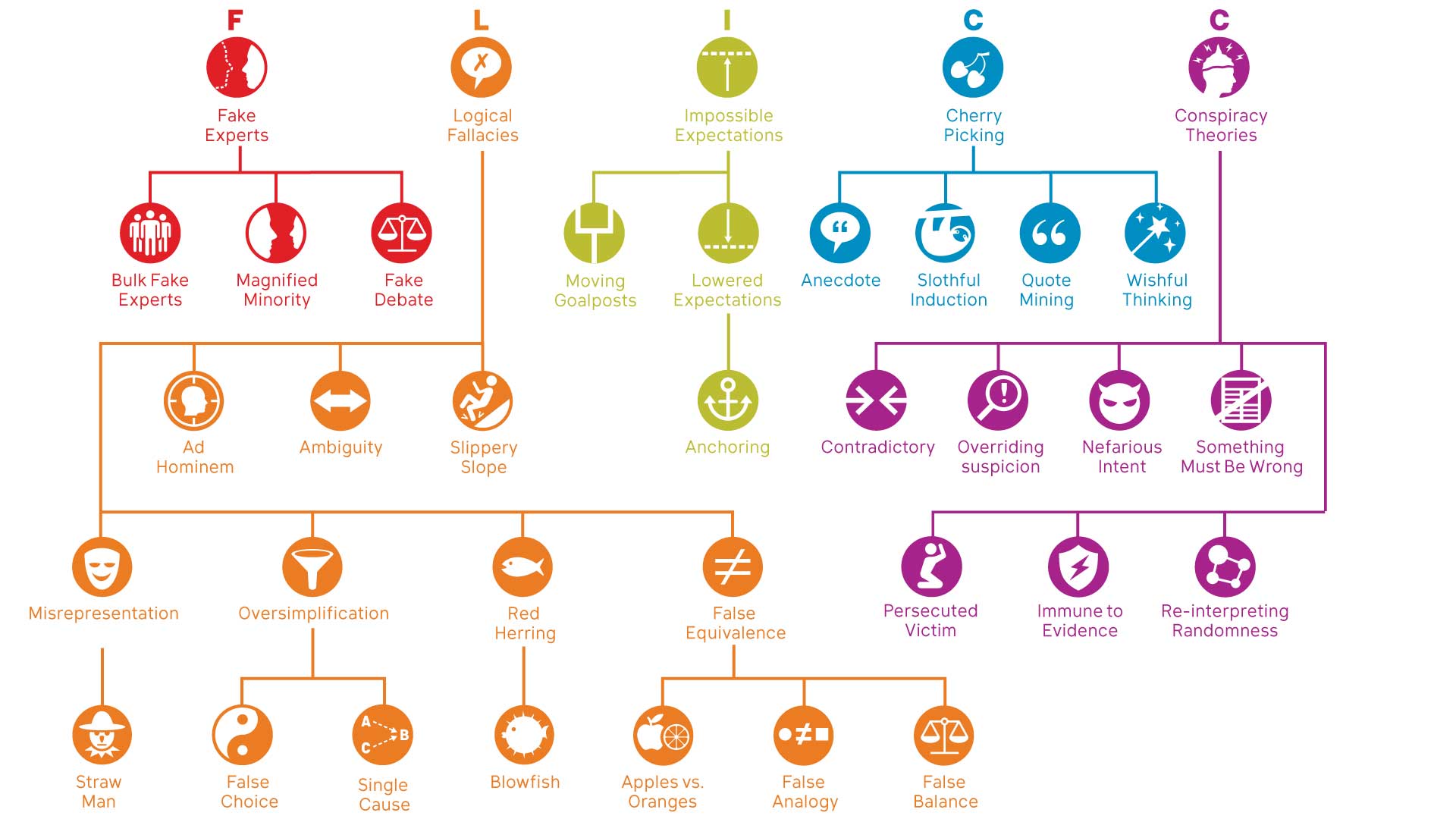}
\end{figure}

Increasing research attention has focused on understanding and countering the techniques used in misinformation. One framework identifies five techniques of science denial---fake experts, logical fallacies, impossible expectations, cherry picking, and conspiracy theories~\citep{diethelm2009denialism}---summarised with the acronym FLICC. These techniques, found in a range of scientific topics such as climate change, evolution, and vaccination, have been developed into a more comprehensive taxonomy shown in Figure~\ref{fig:FLICC}~\citep{cook2020deconstructing}. A critical thinking methodology was developed for deconstructing and analysing climate misinformation in order to identify misleading logical fallacies~\citep{cook2018deconstructing}. This methodology has been applied to contrarian climate claims in order to identify the fallacies used in specific climate myths~\citep{flack2024deconstruct}. Table~\ref{tab:fallacies} lists the fallacies identified in climate misinformation, as well as their definitions. The two types of fallacies are structural, where the presence of the fallacy can be gleaned from the structure of the text, and background knowledge, where certain factual knowledge is required in order to perceive that the argument is fallacious. Table~\ref{tab:fallacies} also presents the logical structure of each fallacious argument.

\begin{table}[h]
\centering
\caption{\label{tab:fallacies}Fallacy types, definitions, and argument structure.}
\rowcolors{2}{gray!10}{white}
\small{\centering\setlength{\tabcolsep}{1mm}\renewcommand{\arraystretch}{1.05}\setstretch{0.96}
\begin{tabular}{  m{2.5cm}  m{1.7cm}  m{5cm} m{3.5cm} } 
      \toprule
      Fallacy & Type & Definition & Argument Structure \\ 
      \midrule
      Ad hominem & Structural & Attacking a person/group instead of addressing their arguments & A has a negative trait. Therefore, A is not credible. \\ 
      
      Anecdote & Structural & Using personal experience or isolated examples instead of sound arguments or compelling evidence & Y occurred once with X. Therefore, Y will occur every time with X. \\ 
      
      Cherry Picking & Structural & Selecting data that appear to confirm one position while ignoring other data that contradicts that position & Group A are lying to us to implement a secret plan. \\ 
      
      Conspiracy\hspace*{0.8mm}theory\hspace*{-3mm}& Structural & Proposing that a secret plan exists to implement a nefarious scheme such as hiding a truth & A is true. B is why the truth cannot be proven. Therefore, A is true. \\ 
      
      Fake experts & Structural & Presenting an unqualified person or institution as a source of credible information. & P has expertise in a non-climate topic. Therefore, P is an expert on climate.\\ 
      
      False choice & Structural & Presenting two options as the only possibilities, when other possibilities exist & P or Q. P. Therefore, not Q. \\ 
      
      False equivalence & Structural & Incorrectly claiming that two things are equivalent, despite the fact that there are notable differences between them. & A and B both share characteristic C. Therefore, A and B share some other characteristic D. \\ 
      
      Impossible expectations & Structural & Demanding unrealistic standards of certainty before acting on the science & There is not enough data or research about X to understand X properly. \\ 
      
      Misrepresentation & Background knowledge & Misrepresenting a situation or an opponent’s position in such a way as to distort understanding &  \\ 
      
      Oversimplification & Background knowledge & Simplifying a situation in such a way as to distort understanding, leading to erroneous conclusions &  \\ 
      
      Single cause & Structural & Assuming a single cause or reason when there might be multiple causes or reasons & X caused Y; therefore, X was the only cause of Y. \\ 
      
      Slothful\hspace*{0.8mm}induction\hspace*{-3mm}& Background knowledge & Ignoring relevant evidence when coming to a conclusion &  \\ 
      \bottomrule
    \end{tabular}}

\end{table}

While these theoretical frameworks have been developed based on psychological and critical thinking research, developing practical solutions countering misinformation is challenging for various reasons. The public perceives misinformation as more novel than factual information, resulting in it spreading faster and farther through social networks than true news~\citep{vosoughi2018spread}. Further, once people accept a piece of misinformation, they continue to be influenced by it even if they remember a retraction, a phenomenon known as the continued influence effect \citep{ecker2010explicit}. To address these challenges, research has begun to focus on pre-emptive or rapid response solutions such as inoculation or misconception-based learning \citep{cook2017understanding}.

One proposed solution is automatic and instantaneous detection and fact-checking of misinformation, described as the ``holy grail of fact-checking''~\citep{hassan2015quest}. Machine learning models offer a tool towards achieving this goal. For example, topic analysis offers the ability to analyse large datasets with unsupervised models that can identify key themes. This approach has been applied to conservative think-tank (CTT) websites, a prolific source of climate misinformation~\citep{boussalis2016text}. Similarly, topic modelling has been combined with network analysis to find an association between corporate funding and polarizing climate text~\citep{farrell2016corporate}. Lastly, topic modelling of newspaper articles has been used to identify economic or uncertainty framing about climate change~\citep{stecula2019framing}. While the unsupervised approach offers general insights about the nature of climate misinformation with large datasets, it does not facilitate detection of specific misinformation claims which is necessary in order to generate automated fact-checks.

\begin{figure}[h]
\caption{\label{fig:CARDStaxonomy}CARDS taxonomy of contrarian climate claims~\citep{coan2021computer}.}
\centering
\includegraphics[width=0.8\linewidth]{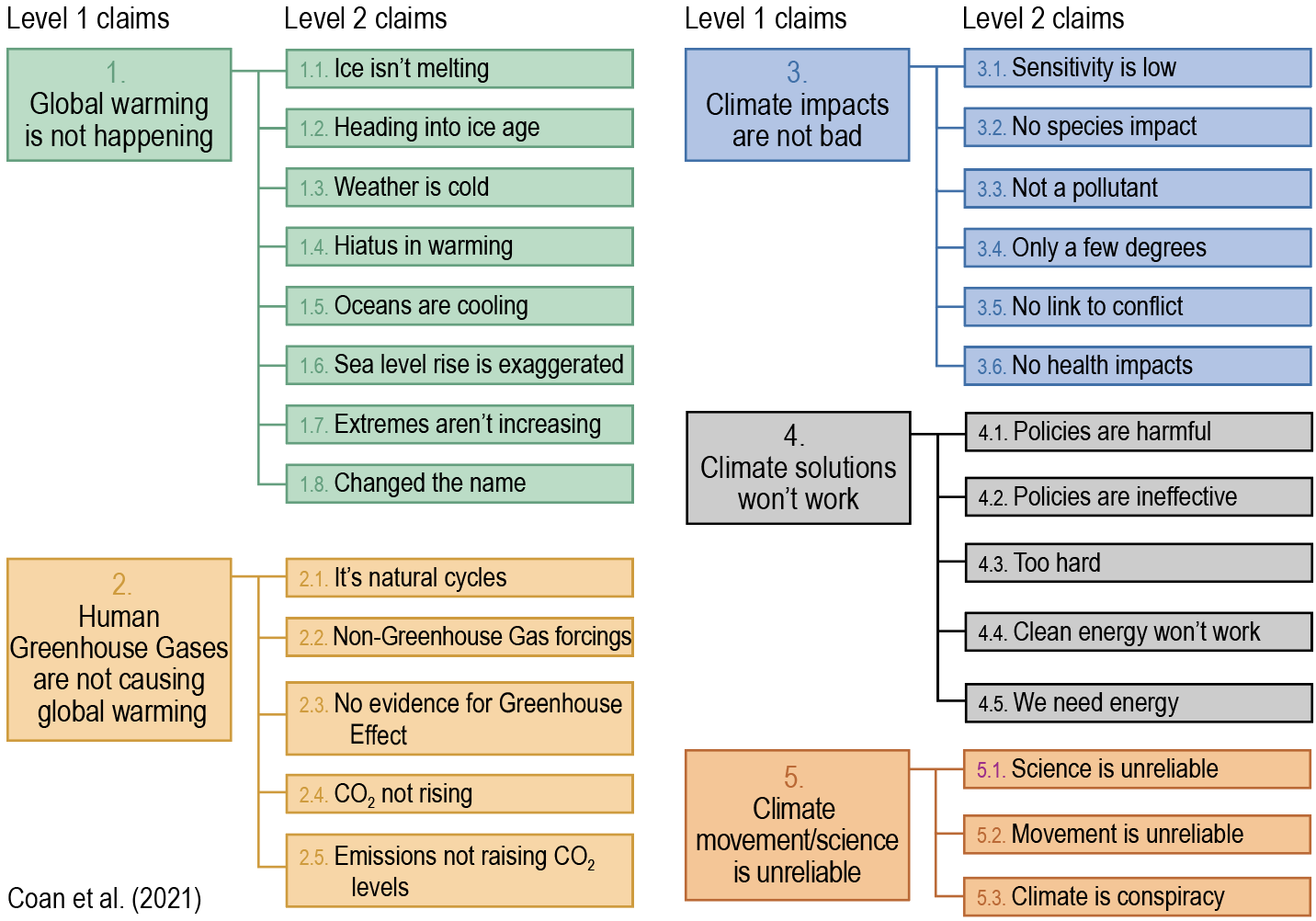}
\end{figure}

To address this shortcoming, a supervised machine model---the CARDS model (Computer Assisted Recognition of Denial and Skepticism)---was trained to detect specific contrarian claims about climate change~\citep{coan2021computer}. To achieve this, the CARDS taxonomy was developed, organising contrarian claims about climate change into hierarchical categories (see Figure~\ref{fig:CARDStaxonomy}). In contrast to the technique-based FLICC taxonomy, the CARDS taxonomy takes a fact-based approach, examining the content claims in contrarian arguments. The CARDS model has been found to be successful in detecting specific content claims in contrarian blogs and conservative think-tank articles~\citep{coan2021computer} as well as in climate tweets~\citep{rojas2024twitter}.

While the CARDS model was developed in order to facilitate automatic debunking of climate misinformation, it by design was only able to detect content-claims. \citet{flack2024deconstruct} found that contrarian claims in the CARDS taxonomy often contained multiple logical fallacies. As an effective debunking needs to contain both explanation of the facts and the fallacies employed by the misinformation\citep{lewandowsky2020debunking}, automated detection of climate misinformation needs to include not only content-claim detection such as that provided by the CARDS model but also detect any fallacies contained in the misinformation.

Several studies have utilized machine learning to detect logical fallacies in climate-themed text. \citet{jin2022logical} developed a structure-aware model to detect fallacies in both climate text and general text, emphasising the importance of the argument's form or structure over its content words.  However, certain fallacies, as indicated in Table~\ref{tab:fallacies}, do not strictly adhere to a fixed structure, requiring a background knowledge base for detection. Alternatively, \citet{alhindi2023multitask} employed instruction-based prompting to detect 28 fallacies across a range of topics, including climate change. Despite these efforts, past studies have demonstrated low accuracy in fallacy detection, and the frameworks used showed limited overlap with FLICC and CARDS frameworks specifically developed for climate misinformation detection and debunking. After closely examining the datasets from~\citet{jin2022logical} and~\citet{alhindi2023multitask}, which are available at \footnote{\url{https://github.com/causalNLP/logical-fallacy}} and \footnote{\url{https://github.com/Tariq60/fallacy-detection}}, we found several data quality issues. These issues included duplicate samples, instances of duplicate samples with different labels, sample repetition across training, validation, and test sets, label merging, empty samples, and ultimately, discrepancies between our formulated fallacy definitions and their annotations.

Our study integrated past psychological, critical thinking, and computer science research in order to develop a technocognitive solution to fallacy detection. Technocognition is the synthesis of psychological and technological research in order to develop holistic, interdisciplinary solutions to misinformation~\citep{lewandowsky2017beyond}. By synthesising the CARDS and FLICC framework, we developed an interdisciplinary solution to fallacy detection that could subsequently be implemented in automated debunking solutions, bringing this research closer to the ``holy grail of fact-checking''.



\section{Results}\label{sec2}

\subsection{Baseline}

The initial step involved establishing a ZeroR classifier, i.e., a classifier that always selects the most frequent class. Our test set comprised a stratified random sampling, where the most frequent label is ``Ad Hominem'', occurring 37 times out of 256 instances. We present the derived accuracy of 0.14 and $F_{1}$ scores of 0.02. These scores can be calculated by employing the respective formula \ref{acc_score} for the accuracy score and \ref{F1_score} for the $F_{1}$ score where TP is the number of true positives, TN is the number of true negatives, FN is the number of false negatives, and FP is the number of false positives.

\begin{equation}
    \label{acc_score}
    Accuracy = \frac{TP+TN}{TP+TN+FP+FN}
\end{equation}
\newline
\begin{equation}
    \label{F1_score}
    F_1 = \frac{2*TP}{2*TP+FP+FN}
\end{equation}

\subsubsection{Comparing our model to Google's Gemini and OpenAi's GPT-4}

Assessing the reasoning skills of large language models (LLMs) is an active area of research, where natural language inference is one of their hardest tasks. One of our goals was to compare our tool to LLMs by applying our test set of 256 samples to Google's Gemini (Gemini-1.0-pro)~\citep{geminiteam2024gemini} and OpenAI's GPT-4 (GPT-4-0125-preview)~\citep{openai2024gpt4} using their respective APIs. We used the following prompt: ``Please classify a piece of text into the following categories of logical fallacies: [a list of all logical fallacy types]. Text: [Input text] Label: ''

The overall accuracy scores for Gemini-pro and GPT-4 in detecting labels were 0.21 and 0.32, both surpassing the ZeroR classifier by 1.5 and 2.3 times. Although LLMs showed an improvement over the most simple baseline, still far from being a reliable tool for this task. In a detailed analysis of these results, Gemini-pro failed to label eight out of the 256 samples with empty responses or replying "None of the above". Gemini-pro's most common predictions were "Oversimplification" (158), "Conspiracy theory" (45) and "Cherry picking" (20). Also, the safety settings were disabled in order to obtain Gemini-pro predictions, as some myths were blocked by the API.

GPT-4, on the other hand, failed to label  44 out of the 256 samples by providing unrequested information and comments such as "... the closest interpretation could be cherry picking" or "The provided text does not seem to fall into any of the listed categories ... Label: None". In these cases, the most likely label was assigned so that in the examples above, the label would be "cherry picking" and "None." With that consideration, GPT-4 assigned "None" to four samples. Its most frequent predictions were "Oversimplification" (84), "Conspiracy theory" (38) and "Anecdote" (26). Table~\ref{tab:gemini-gpt4} shows the detailed break down of results.

\begin{table}[h]
\centering
\resizebox{0.8\columnwidth}{!}{%
    \begin{talltblr}[
    caption = {Fallacy classification results for Google's Gemini and OpenAi's GPT-4 models. For each class, we report precision (P), recall (R), and $F_1$ score.},
    label = {tab:gemini-gpt4},
    ]{
      cell{1}{2} = {c=3}{c},
      cell{1}{6} = {c=3}{c},
      row{2} = {halign=c},
      hline{2,19} = {-}{0.08em},
      hline{3,15} = {-}{},
    }
                            & Gemini &      &       &  & GPT-4 &      &          &  \\
                            & P       & R    & $F_1$ &  & P     & R    & $F_1$ &  \\
    ad hominem              & 0.00    & 0.00 & 0.00  &  & 0.86  & 0.32 & 0.47     &  \\
    anecdote                & 0.00    & 0.00 & 0.00  &  & 0.46  & 0.50 & 0.48     &  \\
    cherry picking          & 0.45    & 0.29 & 0.35  &  & 0.20  & 0.10 & 0.13     &  \\
    conspiracy theory       & 0.42    & 0.86 & 0.57  &  & 0.53  & 0.91 & 0.67     &  \\
    fake experts            & 0.00    & 0.00 & 0.00  &  & 0.75  & 0.86 & 0.80     &  \\
    false choice            & 0.50    & 0.14 & 0.22  &  & 1.00  & 0.14 & 0.25     &  \\
    false equivalence       & 0.00    & 0.00 & 0.00  &  & 0.20  & 0.12 & 0.15     &  \\
    impossible expectations & 0.00    & 0.00 & 0.00  &  & 0.17  & 0.05 & 0.07     &  \\
    misrepresentation       & 0.14    & 0.09 & 0.11  &  & 0.31  & 0.23 & 0.26     &  \\
    oversimplification      & 0.13    & 1.00 & 0.22  &  & 0.14  & 0.60 & 0.23     &  \\
    single cause            & 0.00    & 0.00 & 0.00  &  & 0.36  & 0.25 & 0.30     &  \\
    slothful induction      & 0.00    & 0.00 & 0.00  &  & 0.12  & 0.08 & 0.10     &  \\
                            &         &      &       &  &       &      &          &  \\
    accuracy                &         &      & 0.20  &  &       &      & 0.32     &  \\
    macro avg               & 0.13    & 0.18 & 0.11  &  & 0.39  & 0.32 & 0.30     &  \\
    weighted avg            & 0.13    & 0.20 & 0.12  &  & 0.40  & 0.32 & 0.31     &  
    \end{talltblr}
}
\end{table}

\subsection{Assessing our model performance at detecting different fallacies}
Table~\ref{tab:f1sum} summarises test $F_{1}$-macro score results for all the analysed models. The poor performance of the Low-Rank Adaptation(LoRa)~\citep{hu2021lora} experiments was surprising. Only \textit{roberta-large} and \textit{bigscience/bloom-560m} succeeded in attaining $F_{1}$-macro scores comparable to those from previous settings. However, neither of these experiments outperformed the previously achieved scores, indicating possible areas for future work.

\definecolor{Gray}{rgb}{0.568,0.568,0.568}
\begin{table}[h]
\centering
\resizebox{\columnwidth}{!}{%
    \begin{talltblr}[
    caption={$F_1$ macro scores, highlighted cells indicate the best model parameter combination for each model. Best model overall was microsoft/deberta-base-v2-xlarge, learning rate 1.0e-5, gamma 4, weight decay 0.01 fine-tuned over 15 epochs.},
    label={tab:f1sum},
    ]{
      row{2} = {c},
      cell{2}{1} = {halign=l},
      cell{1}{2} = {c=3}{c},
      cell{1}{6} = {c=4}{c},
      cell{1}{11} = {c=2}{c},
      cell{1}{14} = {c=2}{c},
      cell{3}{3} = {Gray,fg=white},
      cell{4}{8} = {Gray,fg=white},
      cell{5}{11} = {Gray,fg=white},
      cell{6}{8} = {Gray,fg=white},
      cell{7}{2} = {Gray,fg=white},
      cell{8}{3} = {Gray,fg=white},
      cell{9}{11} = {Gray,fg=white},
      cell{10}{11} = {Gray,fg=white},
      cell{3-10}{2-15} = {halign=c},
      hline{2,11} = {-}{0.08em},
      hline{3} = {-}{},
    }
                                     & Learning rate &         &         &  & Focal loss, gamma param. &      &      &      &  & Weight decay  &      &  & LoRa &      \\
    Model checkpoints                & 1.0E-05       & 5.0E-05 & 1.0E-04 &  & 2                        & 4    & 8    & 12   &  & 0.01          & 0.10 &  & 8    & 16   \\
    bert-base-uncased                & 0.56          & 0.65    & 0.58    &  & 0.64                     & 0.61 & 0.63 & 0.56 &  & 0.64          & 0.62 &  & 0.36 & 0.37 \\
    roberta-large                    & 0.66          & 0.68    & 0.02    &  & 0.01                     & 0.00 & 0.69 & 0.00 &  & 0.01          & 0.00 &  & 0.60 & 0.64 \\
    gpt2                             & 0.42          & 0.56    & 0.47    &  & 0.51                     & 0.45 & 0.46 & 0.46 &  & 0.57          & 0.50 &  & 0.10 & 0.30 \\
    bigscience/bloom-560m            & 0.54          & 0.54    & 0.33    &  & 0.48                     & 0.50 & 0.56 & 0.52 &  & 0.46          & 0.51 &  & 0.44 & 0.44 \\
    facebook/opt-350m                & 0.23          & 0.12    & 0.02    &  & 0.20                     & 0.23 & 0.22 & 0.22 &  & 0.21          & 0.22 &  & 0.07 & 0.07 \\
    EleutherAI/gpt-neo-1.3B          & 0.44          & 0.65    & 0.58    &  & 0.44                     & 0.05 & 0.50 & 0.49 &  & 0.57          & 0.57 &  & 0.33 & 0.33 \\
    microsoft/deberta-base           & 0.67          & 0.63    & 0.62    &  & 0.64                     & 0.63 & 0.65 & 0.56 &  & 0.69          & 0.67 &  & 0.02 & 0.02 \\
    microsoft/deberta-base-v2-xlarge & 0.67          & 0.41    & 0.02    &  & 0.70                     & 0.73 & 0.63 & 0.69 &  & \textbf{0.73} & 0.71 &  & 0.07 & 0.38 
    \end{talltblr}
}
\end{table}

The most effective model overall was microsoft/deberta-base-v2-xlarge~\citep{he2021deberta} with a learning rate of 1.0e-5, focal loss with  gamma penalty of 4, weight decay of 0.01, and fine-tuned by 15 epochs. The detailed breakdown of the results can be found in Table~\ref{tab:class-rep}, with the small gap between validation and test results indicating the model's ability to generalise effectively. Table~\ref{tab:cmatrix} displays the confusion matrix, depicting actual labels on the y-axis and predicted labels on the x-axis. We observed greater $F_{1}$ score performance for fake experts, anecdote, conspiracy theory and ad hominem. In contrast, false equivalence and slothful induction exhibited the lowest $F_{1}$ scores.

\begin{table}[h]
\centering
    \begin{talltblr}[
    caption={FLICC model fallacy classification report. For each class, we report precision (P), recall (R), $F_1$ score for validation and test partitions.},
    label={tab:class-rep},
    ]{
      row{2} = {c},
      cell{1}{2} = {c=3}{c},
      cell{1}{6} = {c=3}{c},
      cell{4-19}{2-8} = {halign=r},
      hline{2,19} = {-}{0.08em},
      hline{3,15} = {-}{},
    }
                            & Validation &      &      &  & Test &      &      \\
                            & P          & R    & $F_1$ &  & P & R    & $F_1$   \\
    ad hominem              & 0.76       & 0.75 & 0.75 &  & 0.81 & 0.78 & 0.79 \\
    anecdote                & 0.95       & 0.86 & 0.90 &  & 0.88 & 0.92 & 0.90 \\
    cherry picking          & 0.69       & 0.66 & 0.67 &  & 0.77 & 0.77 & 0.77 \\
    conspiracy theory       & 0.78       & 0.82 & 0.80 &  & 0.78 & 0.82 & 0.80 \\
    fake experts            & 1.00       & 0.92 & 0.96 &  & 1.00 & 1.00 & 1.00 \\
    false choice            & 0.83       & 0.77 & 0.80 &  & 0.62 & 0.71 & 0.67 \\
    false equivalence       & 0.50       & 0.43 & 0.46 &  & 0.50 & 0.38 & 0.43 \\
    impossible expectations & 0.69       & 0.73 & 0.71 &  & 0.69 & 0.86 & 0.77 \\
    misrepresentation       & 0.63       & 0.63 & 0.63 &  & 0.68 & 0.68 & 0.68 \\
    oversimplification      & 0.88       & 0.58 & 0.70 &  & 0.78 & 0.70 & 0.74 \\
    single cause            & 0.81       & 0.74 & 0.77 &  & 0.81 & 0.66 & 0.72 \\
    slothful induction      & 0.54       & 0.82 & 0.65 &  & 0.50 & 0.56 & 0.53 \\
                            &            &      &      &  &      &      &      \\
    accuracy                &            &      & 0.73 &  &      &      & 0.74 \\
    macro avg               & 0.75       & 0.73 & 0.73 &  & 0.74 & 0.74 & 0.73 \\
    weighted avg            & 0.75       & 0.73 & 0.73 &  & 0.75 & 0.74 & 0.74 
    \end{talltblr}
\end{table}

\begin{table}[h]
\centering
\resizebox{\linewidth}{!}{%
\begin{talltblr}[caption={Normalised confusion matrix, actual labels on y-axis, predicted labels on x-axis},label={tab:cmatrix},]{
    cell{1}{1} = {r=12}{valign=m,halign=c, cmd=\rot},
    cell{14}{3} = {c=12}{valign=m,halign=c},
    row{13} = {cmd=\rot},
    cell{1-12}{3-14} = {cmd=\mysetcellcolor},
    vline{3,15} = {1-12}{},
    hline{1,13} = {3-15}{},
    cell{13}{3-14} = {valign=h},
    row{1-12} = {ht=0.5cm, halign=c, valign=m},
    column{3-14} = {wd=0.5cm, co=0.0, halign=c, valign=m},
    column{2} = {halign=r},
}
    Actual & ad hominem              & 0.78       &          & 0.03           & 0.11              &              &              &                   & 0.03                    & 0.03              &                    &              & 0.03               \\
           & anecdote                &            & 0.92     &                &                   &              &              &                   &                         &                   &                    & 0.04         & 0.04               \\
           & cherry picking          & 0.03       &          & 0.77           &                   &              & 0.03         &                   &                         & 0.03              & 0.03               & 0.03         & 0.06               \\
           & conspiracy theory       & 0.14       &          &                & 0.82              &              &              & 0.05              &                         &                   &                    &              &                    \\
           & fake experts            &            &          &                &                   & 1.00         &              &                   &                         &                   &                    &              &                    \\
           & false choice            & 0.14       &          &                &                   &              & 0.71         &                   &                         &                   &                    &              & 0.14               \\
           & false equivalence       & 0.13       &          &                &                   &              &              & 0.38              & 0.25                    &                   &                    & 0.25         &                    \\
           & impossible expectations &            &          &                &                   &              &              &                   & 0.86                    & 0.10              &                    &              & 0.05               \\
           & misrepresentation       &            &          &                & 0.05              &              &              &                   & 0.14                    & 0.68              & 0.09               &              & 0.05               \\
           & oversimplification      &            &          & 0.05           &                   &              &              &                   &                         & 0.05              & 0.70               &              & 0.20               \\
           & single cause            &            & 0.09     & 0.06           &                   &              &              & 0.06              & 0.03                    &                   &                    & 0.66         & 0.09               \\
           & slothful induction      & 0.04       &          & 0.12           &                   &              & 0.08         &                   & 0.04                    & 0.08              & 0.04               & 0.04         & 0.56               \\
           &                         & ad hominem & anecdote & cherry picking & conspiracy theory & fake experts & false choice & false equivalence & imp. expectations & misrepresentation & oversimplification & single cause & slothful induction \\
           &                         & Predicted  &          &                &                   &              &              &                   &                         &                   &                    &              &                    
    \end{talltblr}
}
\end{table}

\subsubsection{Comparing FLICC model to ~\citet{alhindi2023multitask} and ~\citet{jin2022logical}}
Although the comparison is not straightforward, both \citet{jin2022logical} and \citet{alhindi2023multitask} developed climate change fallacy datasets, training machine learning models with similar numbers of fallacies (13 and 9 respectively). They reported overall $F_{1}$ scores of 0.21 and 0.29 for their climate datasets in their best round of experiments, whereas we achieved an $F_{1}$ score 0.73, indicating a performance improvement by a factor of 2.5 to 3.5. However, a direct comparison between these studies and our results are difficult as we do not share the same set of fallacies. But, Table~\ref{tab:alhindi-jin} provides a summary of the results for the shared fallacies between the scores obtained by \citet{jin2022logical} and \citet{alhindi2023multitask} using their respective models on their datasets, and our model's performance on our dataset.

\begin{table}[h]
\centering
\resizebox{0.6\linewidth}{!}{%
\begin{talltblr}[caption={Summary of $F_1$ scores for comparable labels (fallacies). On the left side we have labels from \citet{alhindi2023multitask} and \citet{jin2022logical}. On the right side, the FLICC model labels.},label={tab:alhindi-jin},]{
  column{1} = {r},
  column{2-3} = {c},
  column{4} = {l},
  hline{1,2,5,7,8,11} = {-}{},
}
\citet{alhindi2023multitask} & max. $F_1$ & $F_1$      & FLICC          \\
causal oversimplification                    & 0.53          & \textbf{0.72} & single cause   \\
cherry picking                               & 0.43          & \textbf{0.77} & cherry picking \\
irrelevant authority                         & 0.30          & \textbf{1.00} & fake experts   \\
~                                            & ~             & ~             & ~              \\
                                             &               &               &                \\
\citet{jin2022logical}       & $F_1$     & $F_1$      & FLICC          \\
intentional                                  & 0.25          & \textbf{0.77} & cherry picking \\
ad hominem                                   & 0.42          & \textbf{0.79} & ad hominem     \\
false dilemma                                & 0.17          & \textbf{0.67} & false choice   
\end{talltblr}
}
\end{table}



\section{Discussion}

In this study, we developed a model for classifying logical fallacies in climate misinformation. Our model performed well in classifying a dozen fallacies, showing significant improvement on previous efforts. The Deberta model also showed better results than those obtained from Gemini-pro and GPT-4 models. An interactive tool has been made available online allowing users to enter text and receive model predictions at \url{https://huggingface.co/fzanartu/flicc}.

Nevertheless, our model exhibited lower performance with certain fallacies compared to others, with the false equivalence fallacy displaying the lowest performance, likely due to the relative lack of training examples. However, this factor cannot explain the low performance of slothful induction, which had a relatively high number of training examples. One potential contributor to the difficulty in detecting slothful induction was the conceptual overlap between slothful induction and cherry picking. Both fallacies involve coming to conclusions by ignoring relevant evidence when coming to a conclusion but cherry picking achieves this through an act of commission---citing a narrow piece of evidence that conflicts with the full body of evidence---while slothful induction uses an act of omission---coming to conclusions without citing evidence \citep{flack2024deconstruct}. Another factor to consider in analysing the poor performance of slothful induction as illustrated in Figure~\ref{fig:CARDSvsFallacies} is that the labels of slothful induction and cherry picking stand out as the most widely represented across various topics in CARDS claims. However, cherry picking is concentrated in fewer claims compared to slothful induction, which exhibits is more evenly distributed across all claim topics.

Another source of difficulty are texts that contain multiple fallacies. It is common that climate misinformation incorporates several elements in a single item. An example is making a content claim such as ``a cooling sun will stop global warming'' while also including an ad hominem attack against ``alarmists''. Other research also struggled with the fact that climate misinformation often contains multiple claims, necessitating the need for multi-label classification~\citep{coan2021computer}. Further, some texts may include a single claim that nevertheless contains multiple fallacies. For example, the claim that ``there's no evidence that CO2 drove temperature over the last 400,000 years'' commits slothful induction by ignoring all the evidence for CO2 warming as well as false choice by demanding that either CO2 drives temperature or temperature drives CO2~\citep{flack2024deconstruct}.

Future research could look to improve the model's performance by increasing the number of training examples, particularly for underrepresented fallacies such as false equivalence, fake experts, and false choice. As an active area of research, exploring additional or novel classification models and methodologies, such as LoRa, remains an option. However, our primary interest lies in developing a more comprehensive approach that could potentially bring us closer to the ``holy grail of fact-checking'' a more adept understanding of our deconstructive methodology and imitation of critical thinking within large language models (LLMs). One potentially more accessible avenue involves creating an automated ReAct agent~\citep{yao2023react} that we can further optimise using evolutionary computation techniques, as detailed in~\citep{fernando2023promptbreeder}. A more sustainable, long-term approach might involve fine-tuning a LLM, following the methodologies and findings outlined in \citet{an2023learning} and \citet{huang2023large}.

This study restricted its scope to climate misinformation and fallacies used within contrarian claims about climate change. However, the FLICC taxonomy has also been applied to other topics such as vaccine misinformation~\citep{hopkins2023crankyunclevaccine}. The model could be generalised to tackle general misinformation or other specific topics. Future research could explore combining our fallacy detection model with models that detect contrarian CARDS claims~\citep{coan2021computer, rojas2024twitter}. Potentially, a model that can detect both content claims in climate misinformation and fallacies could generate corrections that adhere to the fact-myth-fallacy structure recommended by psychological research~\citep{lewandowsky2020debunking}.

The issues the model faced with texts that contain multiple fallacies point to an important area of interaction between computer and cognitive science. When misinformation contain multiple fallacies, what is the ideal response from a communication approach? Past analysis has found that climate misinformation frequently contains multiple fallacies~\citep{cook2018deconstructing,flack2024deconstruct}. There is a dearth of  research exploring the optimal communication approach for countering misinformation with multiple fallacies. Figure~\ref{fig:CARDSvsFallacies} illustrates that contrarian climate claims can commit a number of fallacies and as technology to detect these fallacies improves, communication science will need to progress to inform optimal response strategies.

This interaction between psychological and computer science research illustrates the value of the technocognitive approach to misinformation research. Inevitably, technological solutions will interact with humans, at which time psychological factors need to be understood to ensure the interventions are effective. Our model was built from frameworks developed from psychological and critical thinking work~\citep{coan2021computer,cook2018deconstructing,cook2017neutralizing,vraga2020testing}, and any output from such models should be informed by psychological research.

\bibliography{sn-bibliography}


\newpage
\section{Methods}

\subsection{Developing a FLICC/CARDS dataset}

We developed a training dataset that mapped examples of climate misinformation to fallacies from the FLICC taxonomy as well as the contrarian claim in the CARDS taxonomy. Text was manually taken from several datasets: the contrarian blogs and CTT articles in the~\citet{coan2021computer} training set, the climate datasets from~\citet{alhindi2023multitask} and~\citet{jin2022logical}, and the test set of climate tweets from~\citet{rojas2024twitter}. In order to more reliably identify dominant fallacies in text, we employed the critical thinking methodology from~\citet{cook2018deconstructing} to deconstruct difficult examples. Table~\ref{tab:deconstruct} shows a selection of sample deconstructions of the most common combinations of CARDS claims and FLICC fallacies. 

 To further ensure the quality of our manually annotated dataset, we conducted a rigorous examination of our samples. First, we searched for potential duplicates by employing exact matching techniques. Subsequently, we leveraged Bert embeddings~\citep{devlin2019bert} to construct a similarity matrix, utilising cosine similarity (Equation~\ref{cosine_similarity}) as the measure of similarity between samples. We then manually reviewed both the exact matches and pairs of samples with the highest similarity scores and proceeded to remove them. For instance, we identified identical and seemingly identical samples that differed only in extra whitespaces, punctuation marks, or capitalization. We also encountered similar texts referring to distinct records, places, or dates; in such cases, we retained the most representative of these samples. 

\begin{equation}\label{cosine_similarity}
\cos\varphi
  = \frac{\mathbf{A} \cdot \mathbf{B}}{\|\mathbf{A}\|  \|\mathbf{B}\|}
\end{equation}
\newline
\begin{equation}\label{euclidean_distance}
 d\left(p,q\right) = \sqrt{p \cdot p - 2(p \cdot q) + q \cdot q}
\end{equation}

In addition to identifying duplicate samples, we aimed to detect outliers, recognising the possibility of inadvertent misannotation of sample labels. Utilising the same Bert embeddings from before, we calculated the mean embedding for each unique label category. Next, we calculated the Euclidean distance (Equation~\ref{euclidean_distance}) of all samples associated with a particular label from its corresponding mean embedding. We selected 36 samples with notably larger distances. Furthermore, we applied the Isolation Forest algorithm~\citep{IsolationForest}, a robust technique for outlier detection, and identified a set of 50 potential outliers which included the 36 samples identified earlier. Out of these 50 outliers, we did not find misannotated labels, but we selectively removed four samples, primarily for being confusingly worded.

\begin{figure}
\centering
\includegraphics[width=0.90\linewidth]{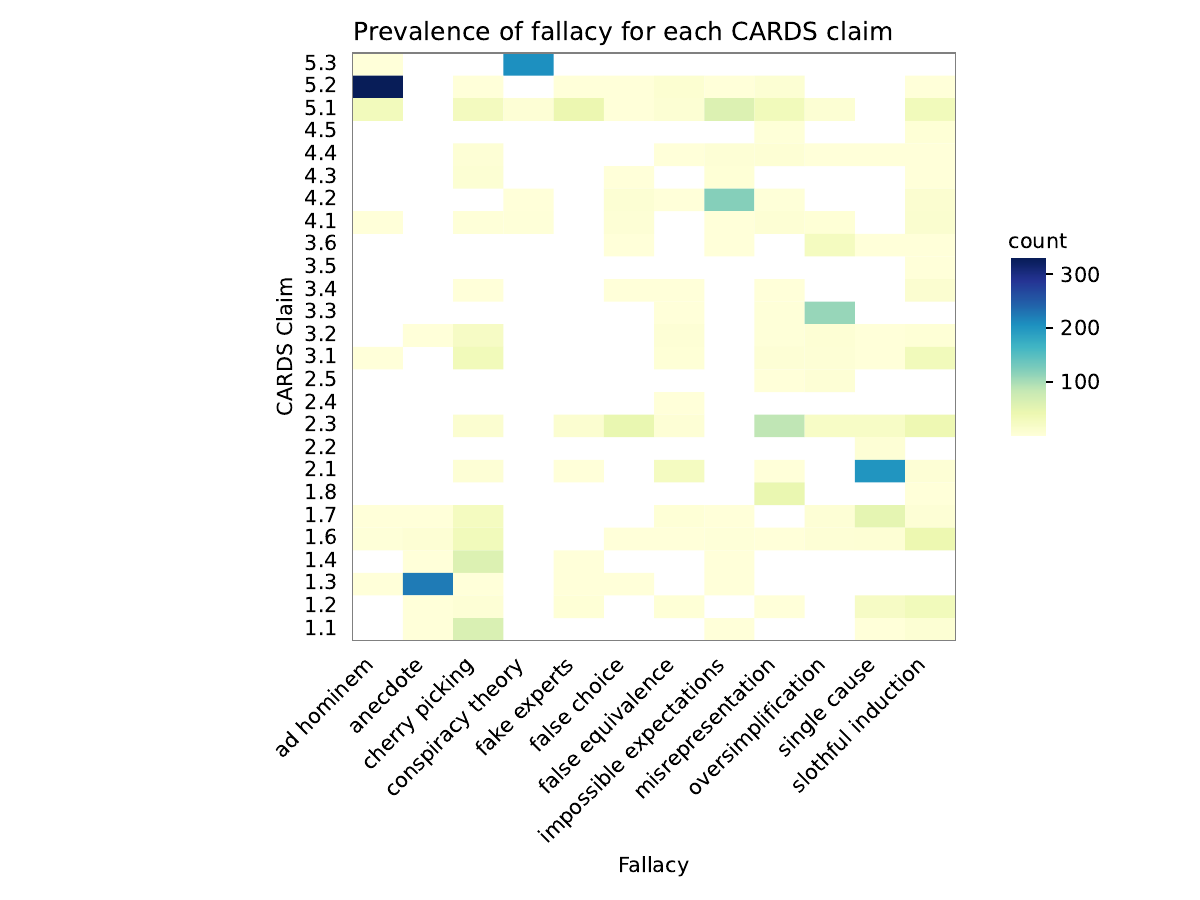}
\caption{\label{fig:CARDSvsFallacies}Map of fallacies across different CARDS claims.}
\end{figure}

The dataset offered a deeper insight into the interplay between FLICC fallacies and CARDS claims, shown in Figure~\ref{fig:CARDSvsFallacies}. It showed a much broader distribution of fallacies within each CARDS claim than found in~\citet{flack2023deconstruct}. This indicated that contrarian arguments could take various forms featuring different fallacies, and that merely detecting a CARDS claim was not sufficient in identifying the argument's fallacy. This underscored the imperative of developing a model for reliably detecting FLICC fallacies in climate misinformation. Our process resulted in a dataset of 2509 samples.

\subsection{Training a Model to Detect Fallacies} 

\subsubsection{Model selection}

Classifying fallacies, especially when they revolve around a singular subject such as climate change, poses a significant challenge. \citet{jin2022logical} contended that this classification task primarily concerned the ``form'' or ``structure'' of the argument rather than the specific content words used. Yet, as depicted in Figure~\ref{fig:CARDSvsFallacies}, it becomes evident that certain fallacies exhibit a higher prevalence within specific claims. 

From the array of available tools, we hypothesised that the low-rank adaptation (LoRa) approach~\citep{hu2021lora} might offer a promising initial solution to our problem. LoRa brings several advantages in terms of storage and hardware efficiency when adapting large language models to downstream tasks. What captivated our interest was how adapting the model weights through trainable rank decomposition matrices could be beneficial for our segmentation problem.

In order to test our hypothesis, we evaluated all accessible models within HuggingFace's Parameter-Efficient Fine-Tuning (PEFT) library \footnote{\url{https://github.com/huggingface/peft}} for sequence classification, with the exclusion of GPT-J due to hardware limitations. Specifically, we tested the following model checkpoints: \textit{bert-base-uncased},\textit{roberta-large}, \textit{gpt2}, \textit{bigscience/bloom-560m}, \textit{facebook/opt-350m}, \textit{EleutherAI/gpt-neo-1.3B}, \textit{microsoft/deberta-base}, \textit{microsoft/deberta-v2-xlarge}.

\subsubsection{Experimental setup}

We employed the PyTorch\footnote{\url{https://pytorch.org}} framework and HuggingFace\footnote{\url{https://huggingface.co}} libraries for our experiments, conducting an iterative analysis to determine the optimal configuration at each experimental stage. Our dataset was partitioned into train, validation, and test sets as illustrated in Table~\ref{tab:datasplits}. The models were trained for a maximum of 30 epochs, and we utilised the validation set to mitigate overfitting by employing an early stopping method after three consecutive rounds without improvement. For each experiment, out of all the training epochs, we selected the model with the best $F_{1}$-macro score, considering the imbalanced nature of our dataset.

\begin{table}[h]
\centering
\begin{talltblr}[caption={Fallacy types and their number of samples on each partition in the FLICC dataset.},label={tab:datasplits},]{
  cell{2-15}{2-6} = {halign=r},
  row{1} = {halign=l},
  hline{1} = {-}{0.08em},
  hline{2,14} = {-}{},
}
Label                   & train & val & test &  & Total \\
ad hominem              & 264   & 67  & 37   &  & 368   \\
anecdote                & 170   & 43  & 24   &  & 237   \\
cherry picking          & 222   & 56  & 31   &  & 309   \\
conspiracy theory       & 154   & 39  & 22   &  & 215   \\
fake experts            & 44    & 12  & 7    &  & 63    \\
false choice            & 48    & 13  & 7    &  & 68    \\
false equivalence       & 52    & 14  & 8    &  & 74    \\
impossible expectations & 144   & 37  & 21   &  & 202   \\
misrepresentation       & 151   & 38  & 22   &  & 211   \\
oversimplification      & 143   & 36  & 20   &  & 199   \\
single cause            & 226   & 57  & 32   &  & 315   \\
slothful induction      & 178   & 45  & 25   &  & 248   \\
Total                   & 1,796 & 457 & 256  &  & 2,509 
\end{talltblr}
\end{table}

We examined the best learning rates within 1.0e-5, 5.0e-5 and 1.0e-4. We set the batch size to 32, employed the AdamW optimiser with a weight decay of 0.0, and utilised the cross-entropy loss function. Once we determined the best learning rate for the model, we moved to the second round of experiments using focal loss~\citep{lin2018focal} instead of cross-entropy loss. Focal loss enables the emphasis on harder-to-classify samples by introducing a gamma penalty to the results; we analysed gamma values of 2, 4, 6, and~16.

Subsequently, we completed a third round of experiments by adding the weight decay parameter, exploring values of 0.1 and 0.01. Again, we did it for the best model identified previously, either with or without focal loss. Finally, we conducted a fourth round of experiments testing LoRa ranks of 8 and 16, as well as alpha values of 8 and~16.

\newgeometry{width=12cm, left=0.6cm, top=0.5cm}
\begin{table}
    \centering
    \small
    \thisfloatpagestyle{empty}
    \caption{\label{tab:deconstruct}Deconstructions of examples of climate misinformation representing 12 fallacies.}
    \begin{tabular}{  m{5cm}  m{.8cm}  m{7cm} m{5.25cm}  }
        \toprule
        Misinformation Example & Claim & Deconstruction & Fallacy Explanation \\ 
        \midrule

         ``I'll believe in climate change when elitists stop building mansions on the coast.'' & 5.2 & P1: Climate advocates argue for climate action.\newline
        P2: Climate advocates' actions are inconsistent with their arguments.\newline
        HP: If climate advocates are inconsistent, their arguments must be invalid.\newline
        C: Arguments for climate action can be disregarded. & HP commits ad hominem, criticising climate advocates rather than their arguments. \\ 
        
          \midrule

         ``Global Warming? Tell that to the southern districts that woke up to negative 10 degrees this morning.'' & 1.3 & P1: Cold weather events are occuring.\newline
        HP: If global warming was happening, we wouldn't experience cold events.\newline
        C: Global warming is not happening. & P1 commits anecdote, using isolated incidents limited in time and place to make conclusions about global warming. \\ 
        
          \midrule
        
         ``Sea ice is setting records this year.'' & 1.1 & P1: In the short term, Arctic sea ice hasn't changed much.\newline
        HP: If Arctic sea ice hasn't changed much in the short term, then it's fine in the long-term.\newline
        C: Arctic sea ice is fine. & HP commits cherry picking, looking at a short period of sea ice data while ignoring the long-term decline in Arctic sea ice. \\ 
        
          \midrule
        
          ``The most extraordinary fraud in the history of Western science: the fantasy that by controlling anthropogenic emissions of carbon dioxide, mankind can control global temperatures.'' & 5.3 & P1: Scientists have commited a range of conspiratorial actions to defend the mainstream view and suppress dissenting views.\newline
        C: There is a conspiracy among scientists to deceive the public. & P1 commits conspiracy theory, assuming that there is secret plotting behind climate science and that scientists act with nefarious intent. \\ 
        
          \midrule
        
         ``More than 31,000 American scientists signed a statement saying they disagree with alarmist predictions.'' & 5.1 & P1: A large number of scientists disagree with human-caused global warming.\newline
        HP: Scientists are experts on climate change regardless of their field of expertise.\newline
        C: There's no scientific consensus on human-caused global warming. & HP commits fake experts. While the signers of the global warming petition project are scientists, almost all of them don't possess climate expertise. \\ 
        
          \midrule
        
         ``Who denies that CO2 lags temperature in the ice core data by as much as 800 years and hence is a product of climate change not a cause?'' & 2.3 & P1: CO2 lagged temperature in the past.\newline
        HP: If temperature affects CO2, then CO2 cannot affect temperature.\newline
        C: CO2 does not drive temperature. & HP presents a false choice between CO2 causing warming or warming causing CO2, while both are true. \\ 
        
          \midrule
        
         ``Tuesday is Earth Day, the calendar's High Holy Day of Green theology. With each passing year, environmentalism more clearly assumes the trappings of a secular religion.'' & 5.2 & P1: The climate change movement have some trait in common with religion.\newline
        HP: A movement that has any traits in common with a religion is a religion.\newline
        C: The climate change movement is a religion. & HP commits false equivalence, making superficial comparisons between the climate movement and religion, when climate science is based on empirical evidence, not faith. \\ 
        
          \midrule
        
         ``A 40\% reduction in US emissions would have no measurable impact on atmospheric CO2 increase.'' & 4.2 & P1: A single policy would have a negligible impact.\newline
        HP: If a single policy doesn't solve global warming, then it is not worth implementing.\newline
        C: We should not have the policy. & HP commits impossible expectations. A single policy cannot solve climate change by itself. We need global cooperation to solve climate change. \\ 
        
            \midrule
        
          ``CO2 is incapable of causing climatic warming. CO2 makes up only 0.038\% of the atmosphere and accounts for only a few percent of the greenhouse gas effect.'' & 2.3 & P1: CO2 is a trace gas, comprising only a small component of the atmosphere.\newline
        HP: If there is a small percentage of CO2 in the atmosphere, its warming potential is low.\newline
        C: CO2 isn't the main cause of global warming. & HP commits misrepresentation as small active substances can have a strong effect (e.g., it only takes a small amount of mercury to poison someone). \\ 
        
          \midrule
        
          ``We, the animals and all land plant life would be healthier if CO2 content were to increase.'' & 3.3 & P1: CO2 is beneficial for plant growth.\newline
        HP: Increased CO2 only has beneficial effects for plants.\newline
        C: Emitting more CO2 will be good for plants. & HP commits oversimplification, ignoring the ways that climate change impacts agriculture through increased heat stress and flooding. The full picture shows that negative impacts outweigh benefits. \\ 
        
            \midrule
        
          ``At the current sea-level-equivalent ice-loss rate of 0.05 millimeters per year, it would take a full millennium to raise global sea level by just 5 cm, and it would take fully 20,000 years to raise it a single meter.'' & 1.6 & P1: Sea level is rising at a modest rate.\newline
        HP: The rate of sea level rise won't increase in the future.\newline
        C: Future sea level rise will not be large. & HP commits slothful induction, ignoring that sea level rise is accelerating and predicted to increase in the future. \\ 
        
            \midrule
        
          ``Yes, there is climate change happening. The world's climate always changes.'' & 2.1 & P1: Climate has changed due to natural causes in the Earth's past.\newline
        P2: Climate is changing now.\newline
        HP: What caused climate change in the past must be the same as what's causing climate change now.\newline
        C: Current climate change must be natural. & HP commits single cause, assuming that what caused climate change in the past (natural factors) must be the same as what's causing climate change now. \\   
  
        \bottomrule
    \end{tabular}
    \label{tab:deconstruct}
\end{table}
\restoregeometry


\end{document}